\documentclass[letterpaper, 10 pt, conference]{ieeeconf}
\IEEEoverridecommandlockouts
\usepackage{amsmath,amssymb,amsfonts}
\usepackage{graphicx}
\usepackage{tikz}
\usepackage{textcomp}
\usepackage{xcolor}
\newcommand{\remove}[1]{}

\usepackage[hidelinks]{hyperref}
\usepackage[style=ieee,backend=biber, url=false, doi=false, isbn=false]{biblatex}
\usepackage{array}
\usepackage{algorithm} 
\usepackage{algpseudocode} 
\AtBeginBibliography{\small}
\newcolumntype{M}[1]{>{\centering\arraybackslash}m{#1}}

\usepackage{graphicx}
\graphicspath{ {./images} }
\def\BibTeX{{\rm B\kern-.05em{\sc i\kern-.025em b}\kern-.08em
    T\kern-.1667em\lower.7ex\hbox{E}\kern-.125emX}}
\newcommand{\deemph}[1]{{\footnotesize{\color{black!80}#1}}}

\newcommand\copyrighttext{%
  \footnotesize \textcopyright 2024 IEEE. Personal use of this material is permitted.
  Permission from IEEE must be obtained for all other uses, in any current or future 
  media, including reprinting/republishing this material for advertising or promotional purposes, creating new collective works, for resale or redistribution to servers or lists, or reuse of any copyrighted component of this work in other works.}
\newcommand\copyrightnotice{%
\begin{tikzpicture}[remember picture,overlay]
\node[anchor=south,yshift=10pt] at (current page.south) {\fbox{\parbox{\dimexpr\textwidth-\fboxsep-\fboxrule\relax}{\copyrighttext}}};
\end{tikzpicture}%
\vspace{-10pt}
}

\begin{document}

\title{\bf{Hierarchical Learned Risk-Aware Planning Framework for Human Driving Modeling}}

\author{Nathan Ludlow$^{1}$ Yiwei Lyu$^{2}$ and John Dolan$^{2}$%
\thanks{$^{1}$Robotics and Dynamics Laboratory, Brigham Young University, Provo, Utah 84602, USA \href{mailto:ndl22@byu.edu}{ndl22@byu.edu}}%
\thanks{$^{2}$CMU Center for Autonomous Vehicle Research, Carnegie Mellon University, Pittsburgh 15289, USA [yiweilyu, jdolan]@andrew.cmu.edu}%
}

\maketitle
\copyrightnotice

\begin{abstract}
This paper presents a novel approach to modeling human driving behavior, designed for use in evaluating autonomous vehicle control systems in a simulation environments. Our methodology leverages a hierarchical forward-looking, risk-aware estimation framework with learned parameters to generate human-like driving trajectories, accommodating multiple driver levels determined by model parameters. This approach is grounded in multimodal trajectory prediction, using a deep neural network with LSTM-based social pooling to predict the trajectories of surrounding vehicles. These trajectories are used to compute forward-looking risk assessments along the ego vehicle's path, guiding its navigation. Our method aims to replicate human driving behaviors by learning parameters that emulate human decision-making during  driving. We ensure that our model exhibits robust generalization capabilities by conducting simulations, employing real-world driving data to validate the accuracy of our approach in modeling human behavior. The results reveal that our model effectively captures human behavior, showcasing its versatility in modeling human drivers in diverse highway scenarios.
\end{abstract}


\section{Introduction}

As the presence of autonomous vehicles on our roads becomes more common, rigorous testing of safety algorithms is essential for the protection of all road users. Real-world testing poses numerous challenges, including cost, time, and safety concerns, particularly in scenarios with potential near-collision events. While real-world validation of systems remains essential, simulations offer a compelling alternative, providing safety, efficiency, cost-effectiveness, and the ability to test and optimize systems under various challenging scenarios without physical risk.

However, a significant challenge in simulation testing stems from the unpredictable nature of human drivers, who often cause near-collision situations involving autonomous vehicles. To ensure accurate and effective simulation testing of autonomous vehicle systems, it's crucial to integrate a precise model of human driving behavior. This ensures that algorithms can adeptly handle real-world unpredictability, ultimately enhancing road safety.

Despite numerous proposed approaches for modeling human behavior, no widely adopted comprehensive model of human driving behavior exists, largely due to limitations in existing models. Human drivers typically plan their actions many seconds in advance, considering numerous driving factors and potential actions of surrounding vehicles, such as lane selection, overtaking decisions, and yielding to oncoming traffic~\cite{lyu2020fg}. To accurately replicate human-like driving behavior for advanced maneuvers, we propose the need to plan the ego vehicle's trajectory several seconds into the future. This planning must account for the multimodal nature of the behavior of surrounding vehicles while optimizing for a safe trajectory, considering all potential surrounding vehicle actions. Many existing human-driving models have limited planning horizons and single-mode surrounding vehicle predictions, which hinder their ability to execute higher-level human behaviors and navigate safely around multiple vehicles. Leveraging an intuitive understanding of how human drivers plan their vehicle trajectories, our proposed model offers improved capabilities in modeling human driving behavior.

To achieve this, we propose a novel hierarchical learned risk-aware planning framework. Our framework uses a recurrent neural network-based trajectory prediction with social pooling method to predict the actions of all surrounding vehicles. Then to capture the interactive risk the ego vehicle faces we propose a matrix to represent the aggregated risk at each point along the highway from a multimodal distribution of possible trajectories from all surrounding vehicles. We use Next Generation Simulation (NGSIM) ~\cite{us_department_of_transportation_federal_highway_administration_next_2017} data to learn parameters that determine human driving and feed those into the risk-aware planning cost function to more accurately represent human driving based on these learned parameters. Since our proposed model is able to plan a long-term trajectory in real-time, this framework is effective at capturing high-level human behaviors, while being sufficiently computationally to be used in simulations with a high number of agents. Additionally, the model includes a parameter for maximum allowable risk, allowing the model to be adjusted to represent different styles of driving, increasing the accuracy of the model in various situations.

Our \textbf{main contributions} are: \textbf{(1)} A novel forward-looking risk metric based on the hierarchical combination of social pooling and history-based trajectory prediction and risk evaluation along predicted trajectories that evaluates risk based not only on current vehicle states, but also predicted future vehicle states that are estimated based on vehicle history and vehicle interaction effects; \textbf{(2)} An accurate human model adjustable with parameters found from data-driven approaches that can represent multiple styles of driving behaviors while being computationally efficient, allowing it to be used in real-time risk assessment and planning of autonomous vehicles. \textbf{(3)} An assessment of the hierarchical learned risk-aware planning framework's performance using real-world data, coupled with simulation-based examinations of its capacity to replicate human driving behavior, accompanied by an analysis of human driver conduct within specific contextual scenarios.

\section{Related Work}

Rule-based approaches attempt to model human driving behavior by intuiting how a human driver will maneuver their vehicle and defining a comprehensive set of rules to represent that behavior. These approaches often generalize well and can easily generate new behaviors not previously observed in datasets. These models also have the benefit of generating easily interpretable trajectories. There are many rule-based approaches for modeling human behaviors such as the Intelligent Driver Model (IDM) ~\cite{treiber_congested_2000}, Game-theoretic planning ~\cite{fisac_hierarchical_2018, yu_human-like_2018}, and risk-aware planning ~\cite{pierson_navigating_2018, pierson_learning_2019}. While the IDM works well in its context, it can only replicate lane following without lane changing, making it unable to replicate complex human behavior. Other approaches can be used to build comprehensive driving models, but fail to  driving behavior that is human-like ~\cite{ahmed_review_2021}.

Learning-based approaches to modeling human driving behavior generally use Inverse Reinforcement Learning ~\cite{sadigh_planning_2016, sadigh_verifying_2019, cao_reinforcement_2020} or Imitation Learning ~\cite{cao_reinforcement_2020, kuefler_imitating_2017, codevilla_exploring_2019} to learn cost functions for human drivers or to replicate expert driving maneuvers, respectively. Both of these methods perform reasonably at replicating human driving behavior when tested in situations already observed in data; however, these methods fail to generalize well to situations and contexts not seen in training data ~\cite{siebinga_human_2022}. Additionally, learning-based methods require large amounts of data for training, which is not always available~\cite{lyu2022adaptive}. Another problem of these learning-based models is that due to the complexity of driving they fail to learn high-level behaviors of vehicles, and only effectively to capture the specific maneuvers of individual vehicles ~\cite{siebinga_human_2022}.

Human driving behavior depends on a number of factors; however, it has been shown that human drivers operate their vehicles in large part based on perceived risk ~\cite{pierson_learning_2019, li_estimation_2017, wang_driving_2015, dingus_driver_2016}. These studies however, do not attempt to use risk modeling to build a comprehensive model of human driving and instead only show that certain high level behaviors can be predicted from the risk the ego vehicle experiences. In our work, we assume that drivers operate their vehicle primarily based on risk~\cite{lyu2023risk,lyu2023decentralized,lyu2023cbf}, and perceived optimality to the driver of a path, and we propose that we can develop an effective driver model with this method. The NGSIM dataset provides valuable information on human driving behavior with proper filtering of data ~\cite{montanino_making_2013, morton_analysis_2017, thiemann_estimating_2008}. By using these data, we observe human driver tendencies and behaviors to learn and validate models on real-world data. Our novel approach to modeling human driving behavior has distinct advantages over existing methods in multiple key areas. By predicting the trajectories of surrounding vehicles multiple seconds into the future, we enable the risk-aware planner to plan for many maneuvers that may be less optimal in the short term but yield long-term rewards, similar to how human drivers often operate their vehicles. Additionally, by learning parameters for the risk-aware planning we can gain many of the benefits of both rule-based and learned models by having a well defined set of driving rules that are generalized to perform well in all road contexts and situations even when the model has not been trained on the specific scenario, while still accurately representing human driving behavior. Lastly, by having adjustable parameters for maximum allowable risk our framework can represent different styles of driving based on the aggressiveness specified for the driver model.

\section{Method}

\subsection{Trajectory Prediction}
Convolutional Social Pooling for trajectory prediction is the means of estimating a set of probable trajectories for a vehicle ~\cite{deo_convolutional_2018}. The trajectory prediction is performed by a set of Long Short-Term Memory (LSTM) networks that are used to encode the histories of surrounding vehicles into a social occupancy grid. These histories are then passed through multiple convolutional and pooling layers before being decoded by another LSTM network into predicted trajectories. The predictor outputs six individual trajectories, representing three lateral maneuvers: left lane change, keep lane, and right lane change; and two longitudinal maneuvers: maintaining velocity or decelerating. Each of these trajectories is assigned an associated probability based on its likelihood of occurrence. The Convolutional Social Pooling approach from ~\cite{deo_convolutional_2018} is employed for its ability to model human driving behavior, its computational efficiency for real-time predictions, and its multimodal output. While Convolutional Social Pooling is well-suited for this purpose, other sufficiently accurate multimodal trajectory prediction methods could be adapted to fit this framework.

\subsection{Risk Measurement} 
Our approach considers all other agents to be dynamic obstacles with respect to the ego agent. We consider a bounded environment $Q \subset \mathbb{R}^{2}$ with points $q \in Q$. The positions of $N_{agents}$ agents in the environment are denoted as $p_j$ with $j = \{1, ...,N_{agents} \}$, and velocities of the agents are denoted with $u_j$. The position of the ego agent is denoted as $p_e$ and the velocity as $u_e$. Each agent is modeled as a kinematic bicycle model with acceleration $a_i$ and steering $\theta_i$ the control input for agent $j$. To account for the uncertain multimodal nature of human driving we use the trajectory prediction module introduced above ~\cite{deo_convolutional_2018} to generate a series of potential trajectories for each actor in the scene at future time steps $t_i$ spaced evenly $t_{s}$ apart where $i = \{1, ...,N_{time}\}$ and $N_{time}$ is the number of predicted vehicle positions. The prediction module can account for interaction effects of other vehicles as well as vehicle histories to provide a more accurate model of potential actions taken by surrounding vehicles. Additionally the trajectory prediction module accounts for multimodality in the potential human actions by generating $N_{traj}$ possible trajectories with the associated probability $P_{ik}$ with $k = \{1, ...,N_{traj} \}$, indicating the normalized likelihood the vehicle will follow the associated trajectory, bounded within the range of $[0, 1]$. Each trajectory is composed of a series of points $p_{ijk}$ spaced $\Delta t$ apart in time, indicating the expected position of the vehicle. This multimodal risk prediction allows us to plan the ego vehicle trajectory based on multiple probable futures for each vehicle similar to how human drivers operate their vehicles.

\begin{equation}
\label{eq_risk_level}
\begin{aligned}
    L_{\bar{P}, i} = \{ q \ \vert \ \mathcal{H}_{c,i}(q,p,\dot{p}) \le \mathcal{H}_{P} \}
\end{aligned}
\end{equation}

For each future time step $t_i$ of the predicted future trajectories of each trajectory we represent the risk at each point $q$ by defining a risk cost as $\mathcal{H}_{c,i}$. As in ~\cite{pierson_learning_2019}, the risk cost represents the risk posed to any point along the road relative to the positions and velocities of all surrounding agents. We define a planning threshold as $\mathcal{H}_{P}$ as the maximum risk the ego vehicle navigate through. This leads to the definition of a risk level set $L_{\Bar{P,i}}$ represented in \autoref{eq_risk_level} at a future time step $t_i$ which is the set of all points less that the chosen planning threshold $\mathcal{H}_{P}$ that the ego vehicle uses when planning.

\begin{figure}[htbp]
\centerline{\includegraphics[scale=0.11, trim={0cm 3.5cm 0cm 2.2cm}, clip]{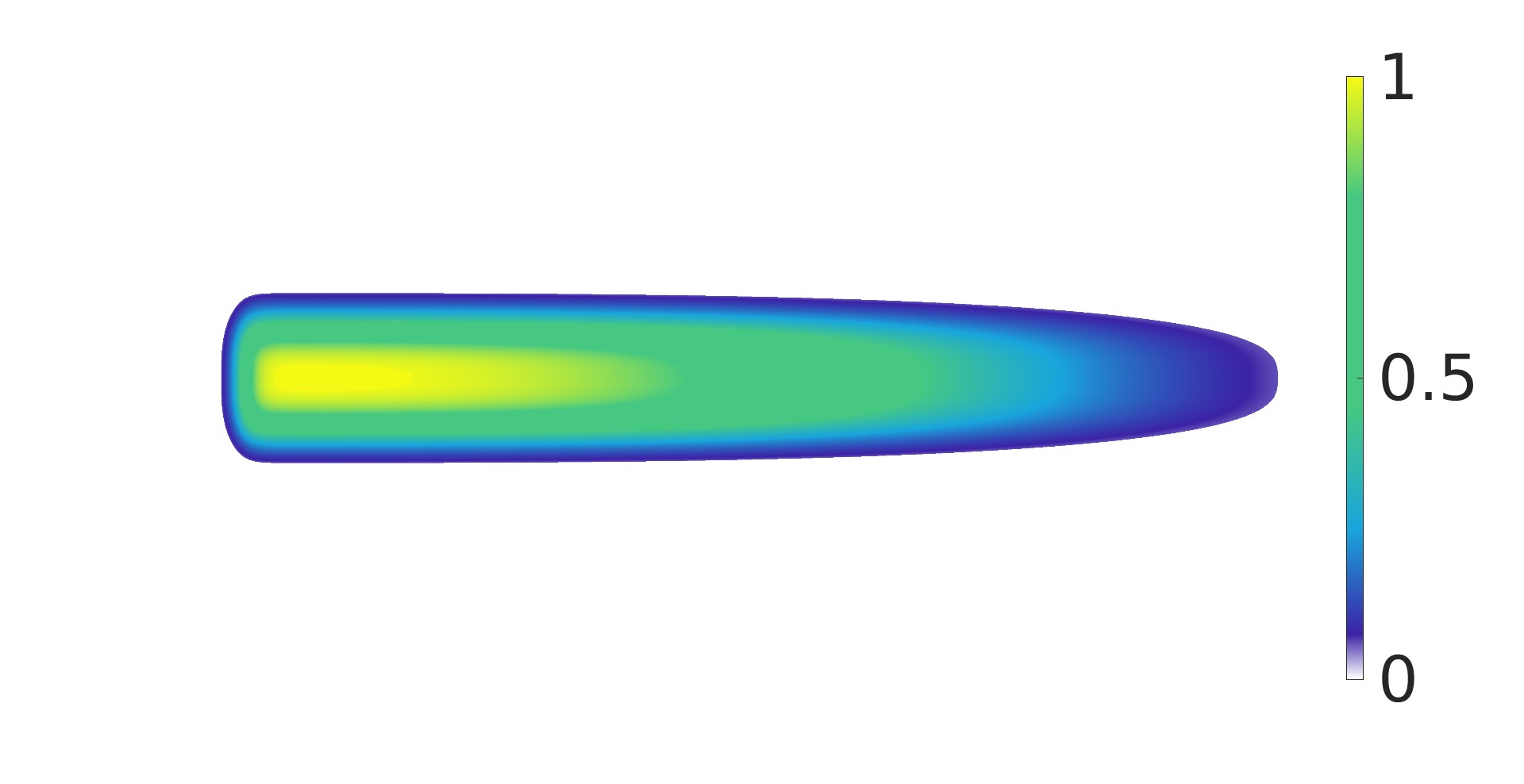}}
\caption{A visualization of the normalized risk computed using the higher-order rectangular Gaussian function for a single vehicle at one time step. Lower risk areas are visualized in blue, and higher risk areas in yellow.}
\label{fig_rectangular_cost}
\end{figure}

\begin{equation}
\label{eq_congestion_cost}
\begin{aligned}
    & \mathcal{H}_{c}(q,p_i,\dot{p_i}) = \\ 
    & \sum^{n}_{i=1}{\dfrac{exp\left(-\left(\dfrac{(q_x - p_{x,i})^2}{\sigma_{x,i}^2}\right)^\beta - \left(\dfrac{(q_x - p_{y,i})^2}{\sigma_{y,i}^2}\right)^\beta\right)}{1 + exp(\alpha \dot{p}_i^T (q - p_i))}}
\end{aligned}
\end{equation}

To compute $\mathcal{H}_{c, i}$ we use rectangular higher-order Gaussians from the work in ~\cite{pierson_learning_2019} represented in \autoref{eq_congestion_cost} with $\sigma_{i} = \dfrac{w_{i}}{2} + |\dot{p}_{i}|$ for agent $i$ with $w_{i}$ representing the size of the agent modeled as a rectangle. We choose to use the rectangular Gaussian as opposed to the elliptical variant, as it weights the risk equally along all points of the body of the agent. This approach allows our method to accurately compute the risk surrounding vehicles of any length or width along the road. Additionally the higher-order Gaussian method of computing risk takes into account both the position and velocity of each agent, meaning that a vehicle with a higher velocity will cause higher risk at a point than an equivalent vehicle with a lower velocity. This intuitively means that as other agent vehicles travel faster the ego vehicle will plan to stay further from them, similar to how human drivers operate their vehicles.

We modify the rectangular cost function to include the probability $P_{jk}$ of the trajectory, since trajectories that are less probable to be taken by the surrounding vehicles cause less risk. As used in ~\cite{pierson_learning_2019}, we use $\alpha = 0.8$ and $\beta = 1.5$, since this provided the best fit to the data based on the behavior from cars in the NGSIM data. By modifying the cost function to take into account the probability of the trajectory being computed we allow the ego vehicle to prioritize avoiding paths that would collide with highly probable trajectories of surrounding vehicles, while also planning to avoid paths that could collide with low likelihood trajectories of surrounding vehicles when there is sufficient space on the road available.

\begin{equation}
\label{eq_risk_cost}
\begin{aligned}
    & \mathcal{H}_{c, i}(q,p_{i},\dot{p_{i}}) = \sum^{N_{agents}}_{j=1} \sum^{N_{traj}}_{k=1} \cdot P_{ij} \cdot \\
    & \dfrac{exp\left(-\left(\dfrac{(q_x - p_{x,ijk})^2}{\sigma_{x,ijk}^2}\right)^\beta - \left(\dfrac{(q_y - p_{y,ijk})^2}{\sigma_{y,ijk}^2}\right)^\beta\right)}{1 + exp(\alpha \dot{p}_{ijk}^T (q - p_{ijk}))}
\end{aligned}
\end{equation}

\autoref{eq_risk_cost} gives us the risk cost for all points $q$ in the planning region at a given future time step $t_i$. We define $\sigma_{ijk} = \dfrac{w_{ijk}}{2} + |\dot{p}_{ijk}|$ for the trajectory $k$ from vehicle $j$ at time step $i$ accounting for the probability of each trajectory and its time step of occurrence. From the risk of all points at the given time we use \autoref{eq_risk_level} to find the risk level set of points the ego vehicle can use to plan for that time step. With this approach we seek to find the risk the ego vehicle will experience at any point along the road over the next few seconds. With this approach we can plan paths that are multiple seconds long before determining which path to take. This is an important aspect of the framework, since human drivers operate their vehicles by thinking about the benefits or dangers or maneuvers over the timeframe of a few seconds. By planning for trajectories that span multiple seconds we can replicate this aspect of human driving and more select paths that imitate human driving behavior.

\subsection{Path Planning and Navigation}

After computing the risk at each future time step $t_i$ that is predicted for each of the other agents we generate a grid of points or nodes spaced $d_p$ along the length of the road $m = \{ 1, ..., N_{horiz} \}$ and for each lane on the road where $l = \{ 1, ..., N_{lanes} \}$ with the position of each node represented as $g_{ml}$. We connect the nodes of this graph with edges that represent possible actions the car can take from any node. These actions are: continue in the current lane, change to the left lane, or change to the right lane. We assign weights $w_{mlc}$ according to \autoref{eq_cost_risk} to each edge with $c = {1, 2, 3}$ for each of the possible lane change actions, and at each node we assign the corresponding level set $L_{\hat{P}, i}$ to each point in the planning grid $L_{ml}$ for if the risk at that point is lower than the risk threshold $\mathcal{H}_{P}$. 

\begin{equation}
\label{eq_cost_risk}
\begin{aligned}
w_{mlc} = \mathcal{H}_{c, i}(g_{ml}, p_{i}, \dot p_{i})
\end{aligned}
\end{equation}

Using the planning grid, we employ Dijkstra's Algorithm to determine the lowest-overall-risk path for the ego vehicle similar to the approach from ~\cite{pierson_learning_2019}. Dijkstra's Algorithm is selected to guarantee that a globally optimal path is found for the ego vehicle, and works efficiently for this algorithm due to the relatively small number of nodes required for effective path for planning. At each point, we verify if $L_{ml}$ falls within the planning level set $L_{\hat{P}, i}$ and incorporate it as a node in the planning graph only when within this level set. If a complete path cannot be established from start to finish, we select the longest path with the lowest cost. This approach minimizes overall risk while avoiding exceeding a maximum safety threshold, thus emulating various human behaviors that prioritize overall safety relative to other vehicles on the road. Thresholding max risk with $L_{P, i}$ prevents the vehicle from taking single overly risky maneuvers that humans would avoid, even if they result in lower risk along the longer path, while optimizing for the lowest-cost total path ensures the vehicle's safety throughout.

\begin{figure}[htbp]
\centerline{\includegraphics[scale=0.19, trim={9cm 14.3cm 12cm 13.6cm}, clip]{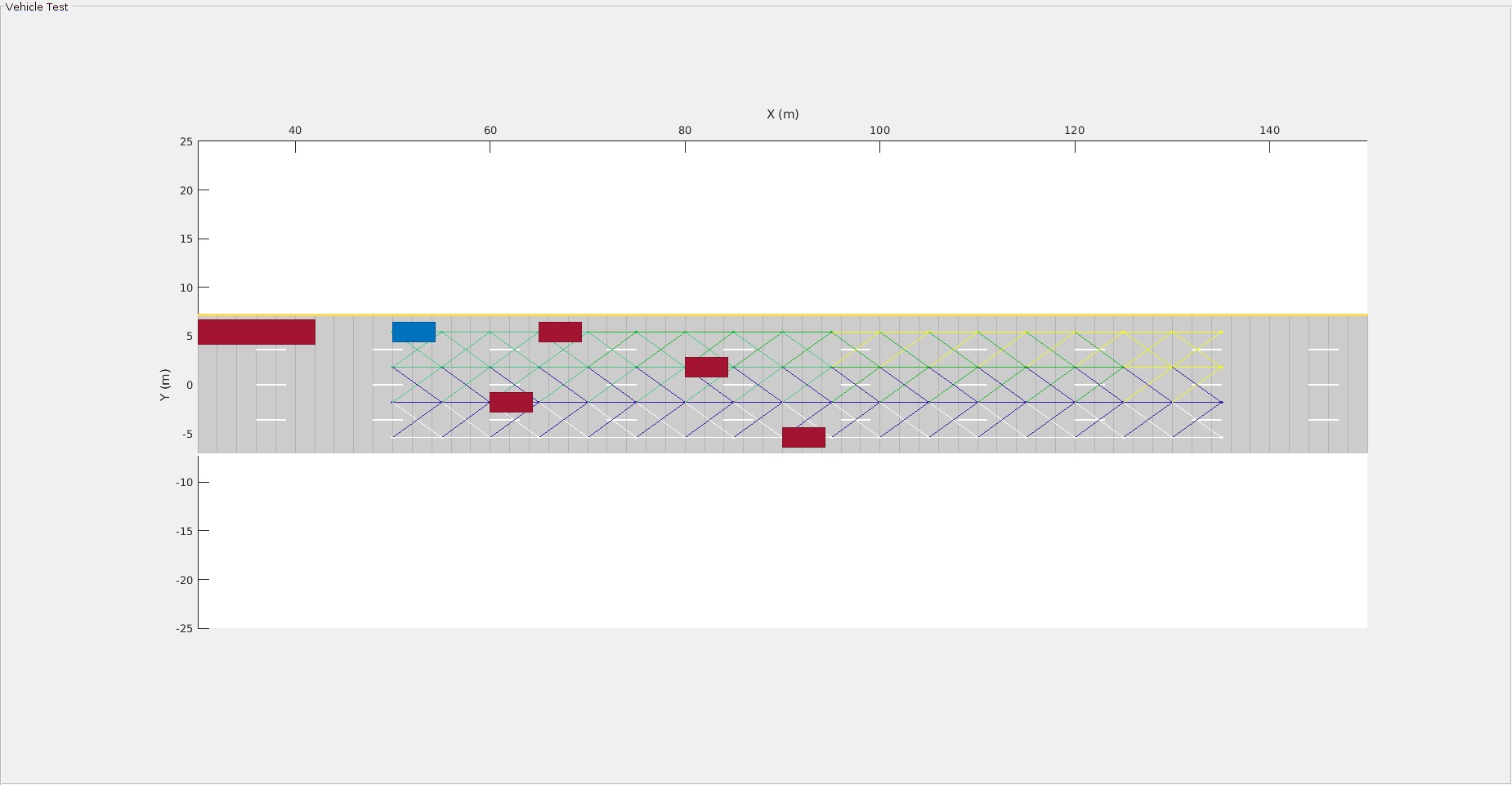}}
\caption{The planning grid with nodes and edges for the ego vehicle (pictured in blue) that is used to find a safe path through the surrounding agents.}
\label{fig_planning_grid}
\end{figure}

\begin{algorithm}\footnotesize
    \caption{Generate Multimodal Risk-Aware Planning Grid} 
    \begin{algorithmic}[1]
        \For {$\forall j \in \{1,...,N_{agents}\}$}
            \State Predict possible agent trajectories ~\cite{deo_convolutional_2018}
        \EndFor
        \State Create graph in planning horizon length
        \For {$\forall i \in \{1,...,N_{time}\}$}
            \State Compute risk $\mathcal{H}_{c, i}$ (\autoref{eq_risk_cost})
        \EndFor
        \For {$\forall m \in \{1,...,N_{horiz}\}$}
            \For {$\forall l \in \{1,...,N_{lanes}\}$}
                \State Find time step $i$ for ego vehicle to reach edge $g_{ml}$
                \State Update $w_{ml}$ for each edge (\autoref{eq_cost_risk})
                \State Set $L_{ml}$ to the risk level set (\autoref{eq_risk_level})
            \EndFor
        \EndFor
    \end{algorithmic} 
    \label{alg_gen_grid}
\end{algorithm}

\subsection{Insights on Learned Parameters}
To more accurately represent human driving behavior as opposed to simply generating safe driving behavior we take the framework previously outlined and learn additional costs that are used when planning trajectories. Not all lanes along a highway travel at the same speed: human drivers prefer to drive in different lanes based on their desired speed of travel. To reflect this behavior in our model, we learn the driver's lane preference based on the current speed of the driver relative to the speed of other vehicles in their lane.

To learn parameters as additional costs for the risk-aware planning, we divide the NGSIM data from I-80 and US-101 in half, reserving half of the data to validate our model. We also filter all data to find the smoothed positions and velocities of all the vehicles to correct errors in the NGSIM data. Finally, we re-extract the current lanes of each vehicle from its position to accurately identify each vehicle's lane at each time step in the NGSIM data. The data used for training the model included approximately 4,500 cars with over 80 hours of combined driving time for learning parameters to model the driving on human behavior.

\begin{table}[htbp]
\caption{Learned Parameters for Model}
\begin{center}
\renewcommand{\arraystretch}{1.5}
\begin{tabular}{ M{1.29cm} M{1.1cm} c c c M{0.9cm} } 
 \hline
 {} & {\bf Lane 1 (Slowest)} & {\bf Lane 2} & {\bf Lane 3} & {\bf Lane 4} & {\bf Lane 5 (Fastest)} \\ 
 \hline
 {\bf 1 $\sigma$ slower} & {0.4224} & {0.5030} & {0.5169} & {0.4994} & {0.4460} \\ 
 {\bf Mean} & {0.4017} & {0.4868} & {0.4862} & {0.4677} & {0.4386} \\ 
 {\bf 1 $\sigma$ faster} & {0.4123} & {0.4962} & {0.5070} & {0.4891} & {0.4421} \\
 \hline
\end{tabular}
\label{tab_learned_parameters}
\end{center}
\end{table}

\begin{figure}[htbp]
\centerline{\includegraphics[scale=0.125, trim={3cm 0.0cm 3.5cm 3.0cm}, clip]{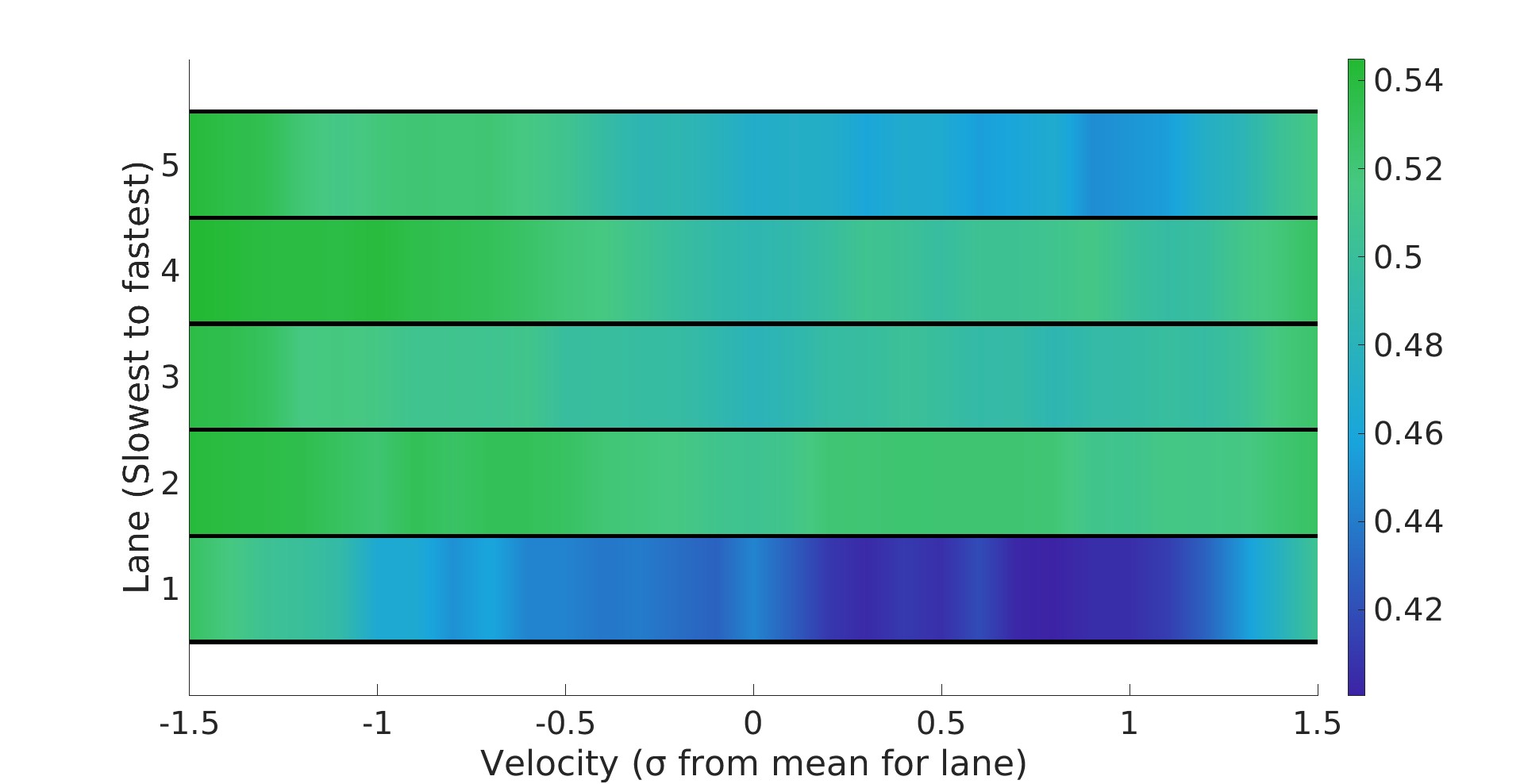}}
\caption{A visualization of the risk in each lane relative to the velocity of the vehicle. This is a visualization of data seen in \autoref{tab_learned_parameters}.}
\label{fig_learned_parameters}
\end{figure}

For each time step of each vehicle in the data, we compute the risk field at the current time step $t_i$ using the rectangular cost function in \autoref{eq_congestion_cost}, and record the risk the vehicle is experiencing, along with its velocity, current lane, and the mean and standard deviation of velocity of all vehicles in the current vehicle's lane. We process these data to determine the average risk a vehicle experiences when traveling within the range of 1.5 standard deviations from the mean velocity of the lane $[-1.5 \sigma, 1.5 \sigma]$. We selected $1.5 \sigma$ since about $90\%$ of all vehicle velocities fall within this range, leading to a good fit for the data without giving too much weight to outliers. The data can be found in \autoref{tab_learned_parameters} (see \autoref{fig_learned_parameters} for a visualization of these data).

To incorporate the learned parameters into our model we modify the weights of the edges of the planning grid with \autoref{eq_total_risk} to reflect the difference in risk that the vehicle would experience by changing lane. We fit a function to the acquired data $\mathcal{H}_{l}(l, u_{avg}, u_{\sigma}, u_{ijk})$ represented by \autoref{eq_delta_h_risk} with an input of the current current lane, average lane velocity, standard deviation of lane velocity and the velocity of a vehicle and output of the average risk in the lane bounded in the range $[0, 1]$. Then we compute the current value of $\Delta \mathcal{H}_{l}$ using \autoref{eq_delta_h_risk}.

\begin{equation}
\label{eq_delta_h_risk}
\begin{aligned}
\Delta \mathcal{H}_{l} = & \mathcal{H}_{l}(l2, u_{l2, avg}, u_{l2, \sigma}, u_{i+1jk}) \\
& - \mathcal{H}_{l}(l1, u_{l1, avg}, u_{l1, \sigma}, u_{ijk})
\end{aligned}
\end{equation}

\begin{equation}
\label{eq_total_risk}
\begin{aligned}
w_{mlc} = \mathcal{H}_{c, i}(g_{ml}, p_{ijk}, \dot p_{ijk}) + \Delta \mathcal{H}_{l}
\end{aligned}
\end{equation}
13:05:31 202
Using \autoref{eq_total_risk} as the new cost function for the edges of the planning grid in Algorithm \ref{alg_gen_grid} generates two new behaviors in the ego vehicle that replicate human driving behavior in highway driving. First, it rewards the ego vehicle for traveling at the speed of the other traffic in the lane. As the ego vehicle speeds up or slows down to reach the same velocity as the other vehicles in the lane the $\Delta \mathcal{H}_{l}$ factor decreases, which lowers the total cost of the path. This behavior of speeding up or slowing down in a lane to match the average speed of other cars in the lane more closely matches human behavior when driving within one lane on a highway as opposed to constantly speeding up and slowing down to maintain a safe distance other vehicles. 

Second, the $\Delta \mathcal{H}_{l}$ factor encourages the ego vehicle to enter lanes that align with the ego vehicle's desired speed. We input the ego vehicle's desired velocity as a planner parameter, which guides it to lanes with velocities matching its preference. When the ego vehicle aims to surpass its current lane's speed, it shifts to a faster lane, lowering its $\Delta \mathcal{H}_{l}$ term. Conversely, it changes to a slower lane when desiring a lower speed than the current lane's average velocity. These emergent behaviors, arising from this single cost term addition, highlight the approach's advantages. Many such terms can be learned from data to improve the modeling of human behaviors in varying driving situations. In the next section, we validate our model's lane change prediction ability using vehicle data and demonstrate its effective use in replicating human driver lane changes. The process for learning parameters can be easily repeated for new parameters, continually generating accurate human behaviors for various driving scenarios.

\section{Simulation and Discussion}

\subsection{Model validation}

In this section, we validate our model using real-world data and show specific examples of our framework generating human-like behavior not found in other simulation methods for human drivers. We validate the model using the approximately 1,000 lane change maneuvers in the half of the NGSIM data that we did not use to learn the parameters. Further, we subdivide the data by examining the lateral velocities of each driver during their maneuvers and categorizing the drivers as conservative, normal, or aggressive. We define drivers with the lower quartile of lateral and longitudinal accelerations during lane changes as conservative, drivers in the upper quartile as aggressive, and the rest as normal. We define three different parameters for aggressiveness that represent the varying driving styles, and validate each lane change maneuver with the corresponding model.

To perform the validation, we replace one of the recorded vehicles from NGSIM data in the simulation with our model 2 seconds before the lane change occurs in the data, and observe the planned trajectory for the next 4 seconds to determine if the model changes lanes into the same lane as the actual data. We implement the previously described method of planning with a model predictive controller and record the final and average deviation of our model from the ground truth. We divide the error into lateral, and longitudinal error to demonstrate the models performance in both directions. We present our results in \autoref{tab_results}.

\begin{table}[htbp]
\caption{Driver Model Validation by Driver Type}
\begin{center}
\renewcommand{\arraystretch}{1.5}
\begin{tabular}{ M{1.60cm} M{1.15cm} M{1.15cm} M{1.15cm} M{1.15cm} } 
 \hline
 & {\bf Aggressive} & {\bf Normal} & {\bf Conservative} & {\bf Total}\\ 
 \hline
 {\bf Accuracy} & {77.41\%} & {81.27\%} & {83.73\%} & {80.92\%} \\
 {\bf RMSE$_{lat, avg}$} & {0.348 m} & {0.312 m} & {0.285 m} & {0.315 m} \\
 {\bf RMSE$_{lat, final}$} & {0.518 m} & {0.436 m} & {0.384 m} & {0.446 m} \\
 {\bf RMSE$_{lon, avg}$} & {3.51 m} & {3.12 m} & {2.98 m} & {3.20 m} \\
 {\bf RMSE$_{lon, final}$} & {5.32 m} & {5.11 m} & {5.01 m } & {5.15 m} \\
 \hline
\end{tabular}
\label{tab_results}
\end{center}
\end{table}

\autoref{tab_results} demonstrates our model's capability to accurately predict lane changes. Notably, prediction accuracy decreases slightly as driver aggressiveness increases. This is likely due to the more variable nature of aggressive driving behaviors. Aggressive drivers often prioritize speed as-opposed-to safety and adherence to traffic rules, expanding the set of potential actions beyond those of conservative drivers. This variability makes it challenging to pinpoint the exact actions of aggressive drivers, as reflected in our results. Importantly, the slight decrease in accuracy for aggressive drivers does not hinder our model's capability to generate plausible aggressive driving maneuvers. The framework may select a slightly different, but realistic aggressive maneuver within the broader range of possibilities.

\begin{table}[htbp]
\caption{Comparison of existing models for predicting lane change maneuvers with 3 seconds of prediction horizon}
\begin{center}
\renewcommand{\arraystretch}{1.5}
\begin{tabular}{ c c c } 
 \hline
 {\bf Model} & {\bf Dataset} & {\bf Accuracy} \\ 
 \hline
 {Proposed} & {NGSIM} & {81\%} \\
 {Support-vector Machine ~\cite{zhang_lane_2022}} & {NGSIM} & {47\%} \\
 {Multi-Layer Perceptrons ~\cite{zhang_lane_2022}} & {NGSIM} & {58\%} \\
 {Random Forest ~\cite{zhang_lane_2022}} & {NGSIM} & {63\%} \\
 {LSTM Network ~\cite{Su-2018-112815}} & {NGSIM} & {86\%} \\
 {Convolutional Neural Network ~\cite{zhang_lane_2022}} & {NGSIM} & {91\%} \\
 \hline
\end{tabular}
\label{tab_lane_change_pred_results}
\end{center}
\end{table}

We provide a comparison of our model's ability to predict lane changes by human drivers in \autoref{tab_lane_change_pred_results}. While the Convolutional Neural Network and LSTM Network yield slightly better results in lane prediction than our proposed model, we assert that our model remains valid as a method for modeling human driving behavior as our model prioritizes generating new human-like trajectories over precise lane change predictions. Furthermore, we comprehensively model the entire highway driving scenario, providing insights into both the likelihood and location of lane changes, in contrast to the binary classification of lane change occurrence offered by many other methods. We additionally emphasize the uniqueness of our approach in modeling human driving behavior based on adjustable hyperparameters for aggressiveness. These parameters enable our model to generate multiple distinct human-like trajectories for the same initial highway scenario leading to an improved capability of our model to be used for testing autonomous vehicle frameworks.

We acknowledge that training exclusively on NGSIM data restricts the model's ability to reproduce driving maneuvers absent from the dataset. However, NGSIM's diverse range of drivers suggests that, with sufficient data on other driving maneuvers, this approach can learn additional behaviors.

\subsection{Example Cases}
To demonstrate the ability of this model to generate new human behaviors not seen in data we present the following scenarios which showcase our framework's ability to generate diverse behaviors. All scenarios shown below are sequential snapshots of egocentric views of a simulated highway environment with the ego vehicle is depicted in dark blue and the planned trajectory shown by the pink line.

\subsubsection{Weaving through traffic}

\begin{figure}[htbp]

\deemph{t = 0.0s} \\
\centerline{\includegraphics[scale=0.435, trim={25cm 8.1cm 21cm 23cm}, clip]{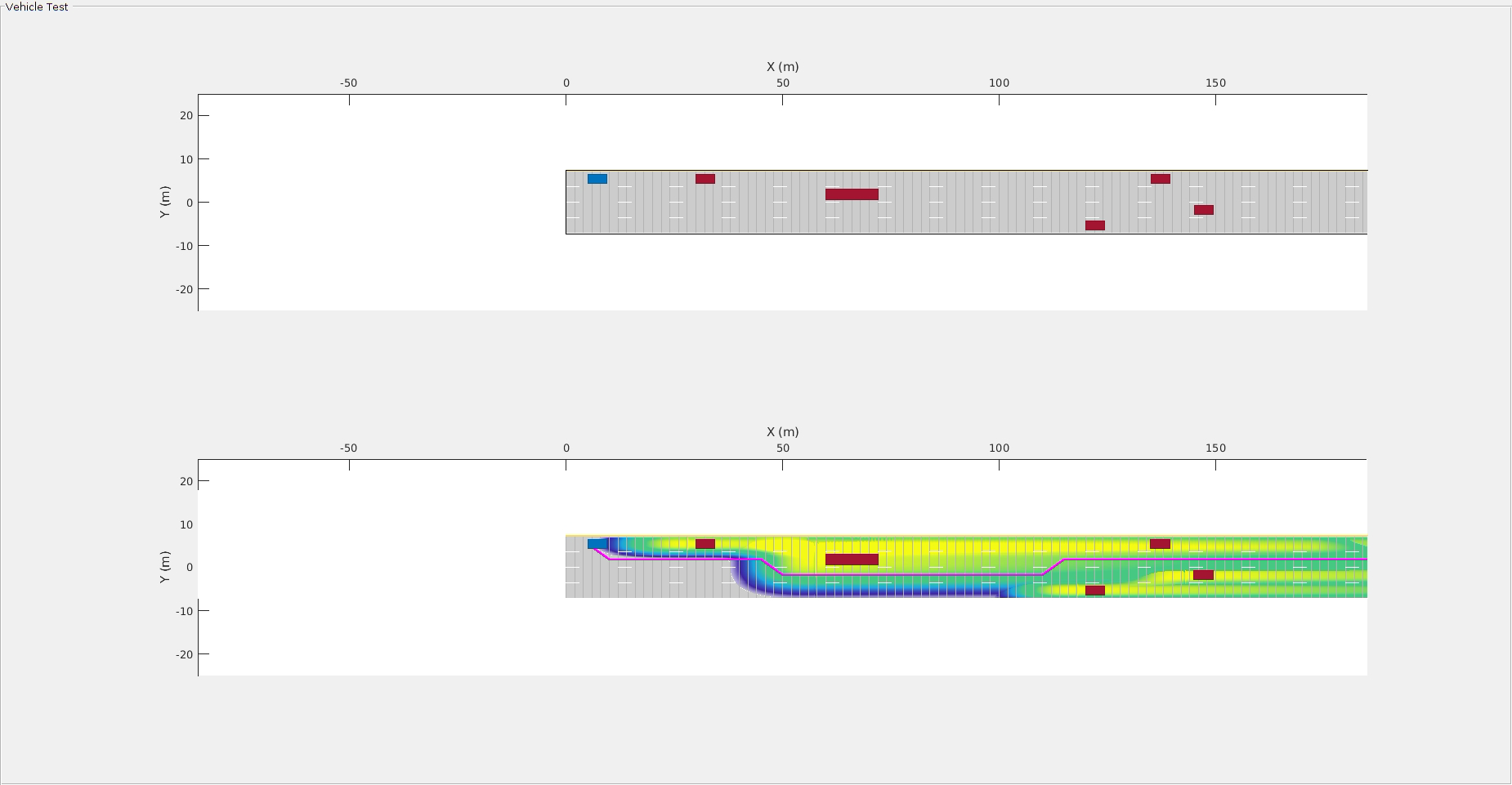}}
\deemph{t = 0.7s} \\
\centerline{\includegraphics[scale=0.435, trim={25cm 8.1cm 21cm 23cm}, clip]{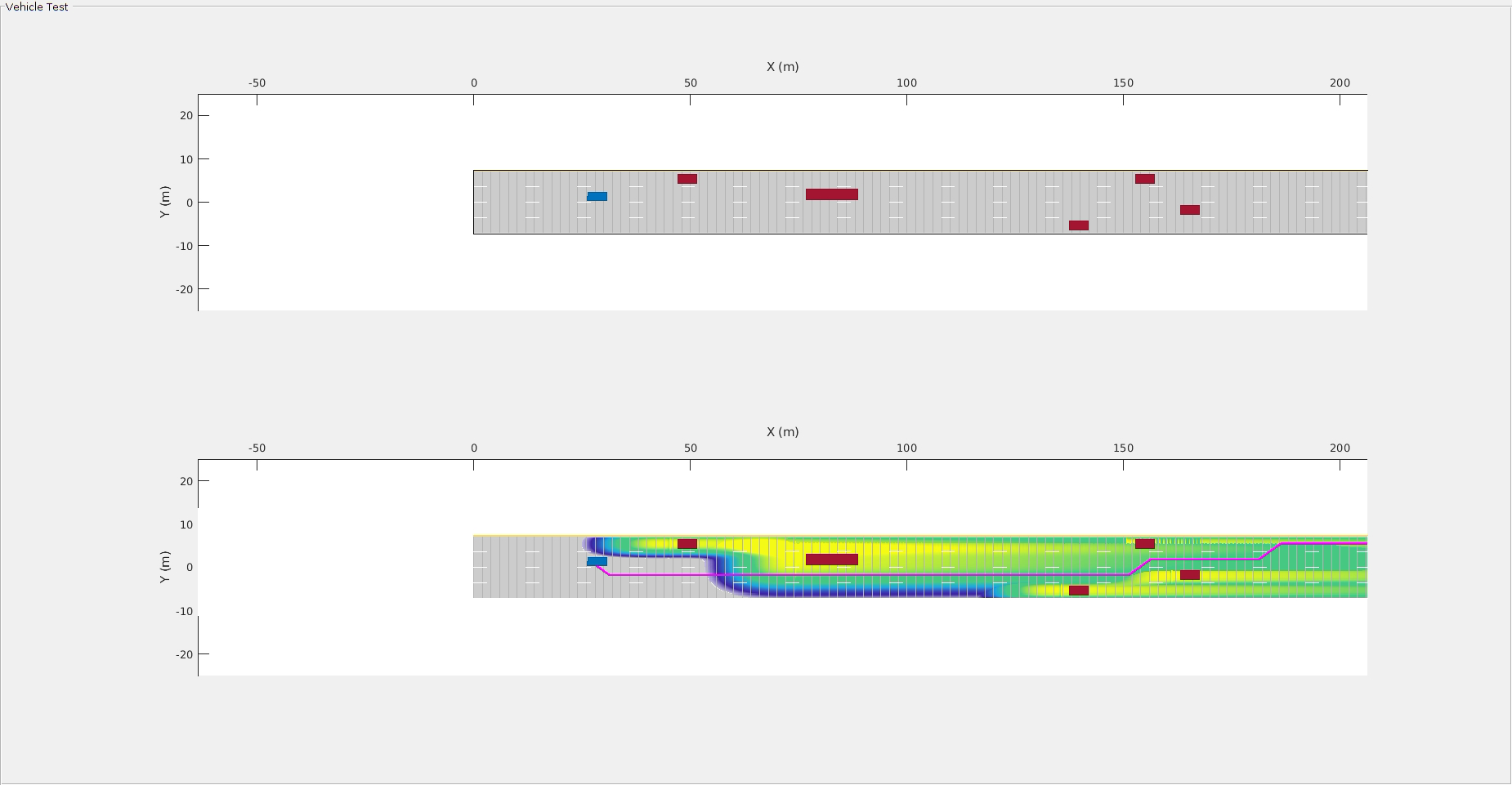}}
\deemph{t = 5.6s} \\
\centerline{\includegraphics[scale=0.435, trim={25cm 8.1cm 21cm 23cm}, clip]{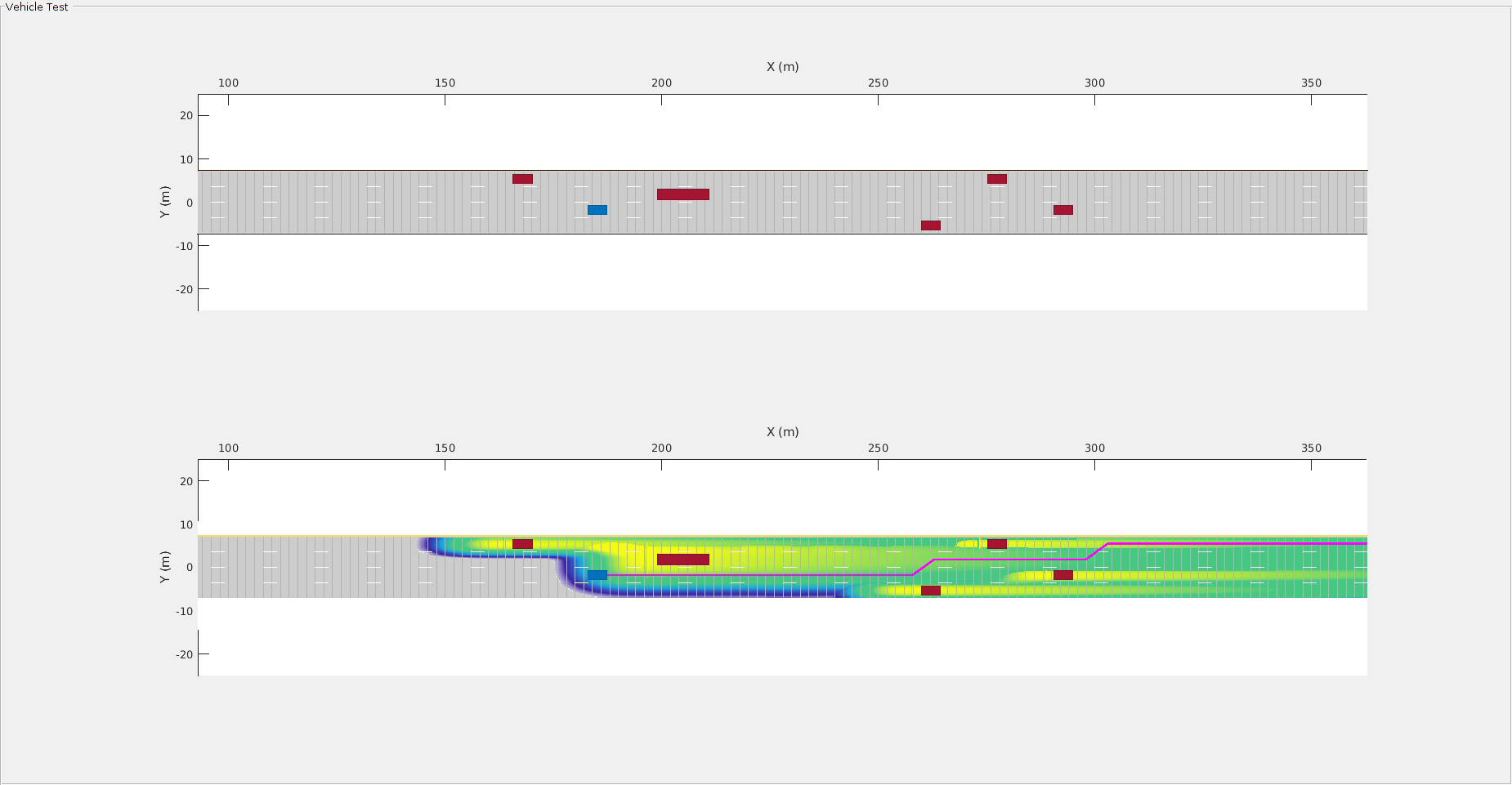}}
\deemph{t = 15.7s} \\
\centerline{\includegraphics[scale=0.435, trim={25cm 8.1cm 21cm 23cm}, clip]{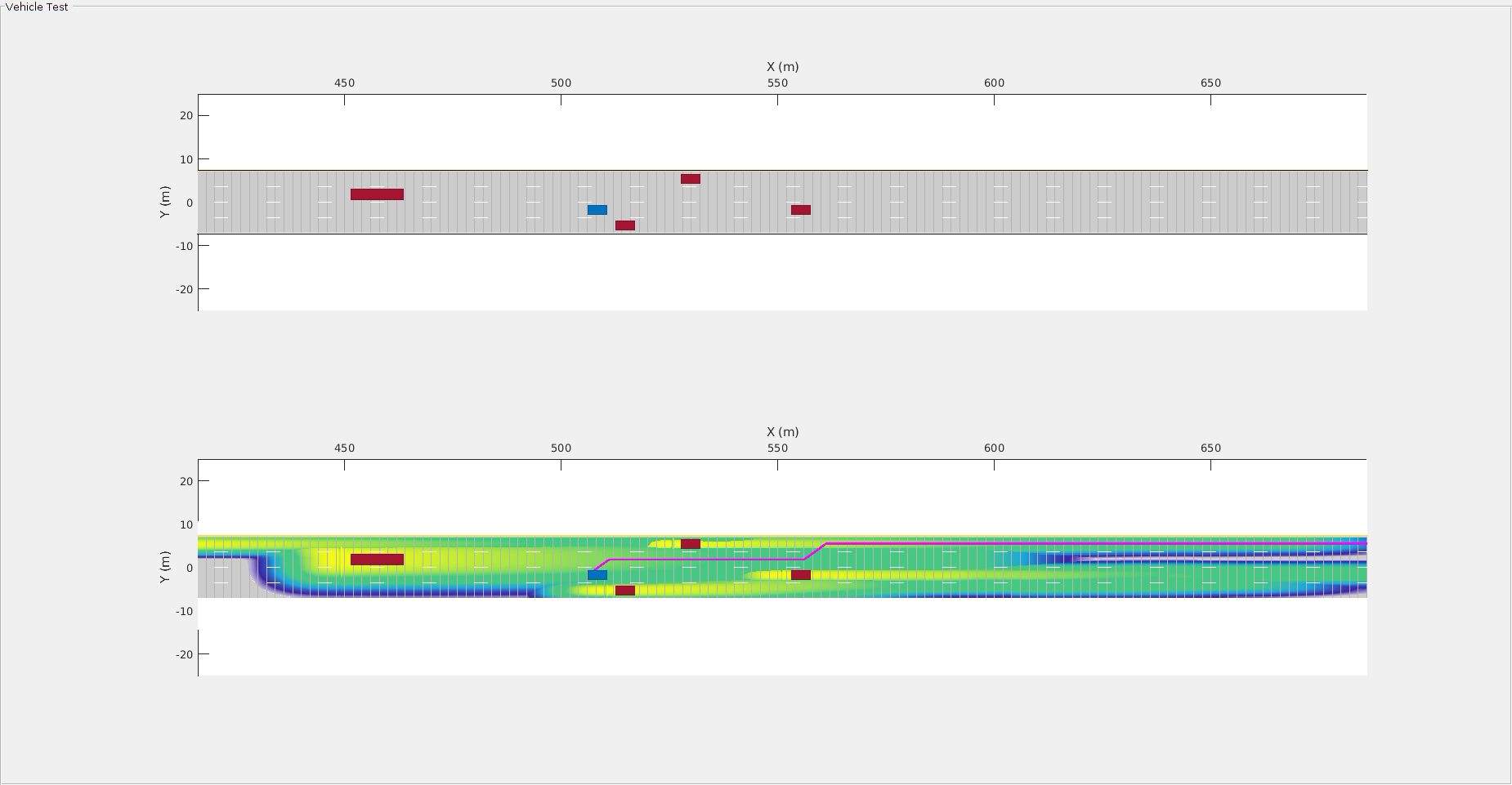}}
\deemph{t = 17.7s} \\
\centerline{\includegraphics[scale=0.435, trim={25cm 8.1cm 21cm 23cm}, clip]{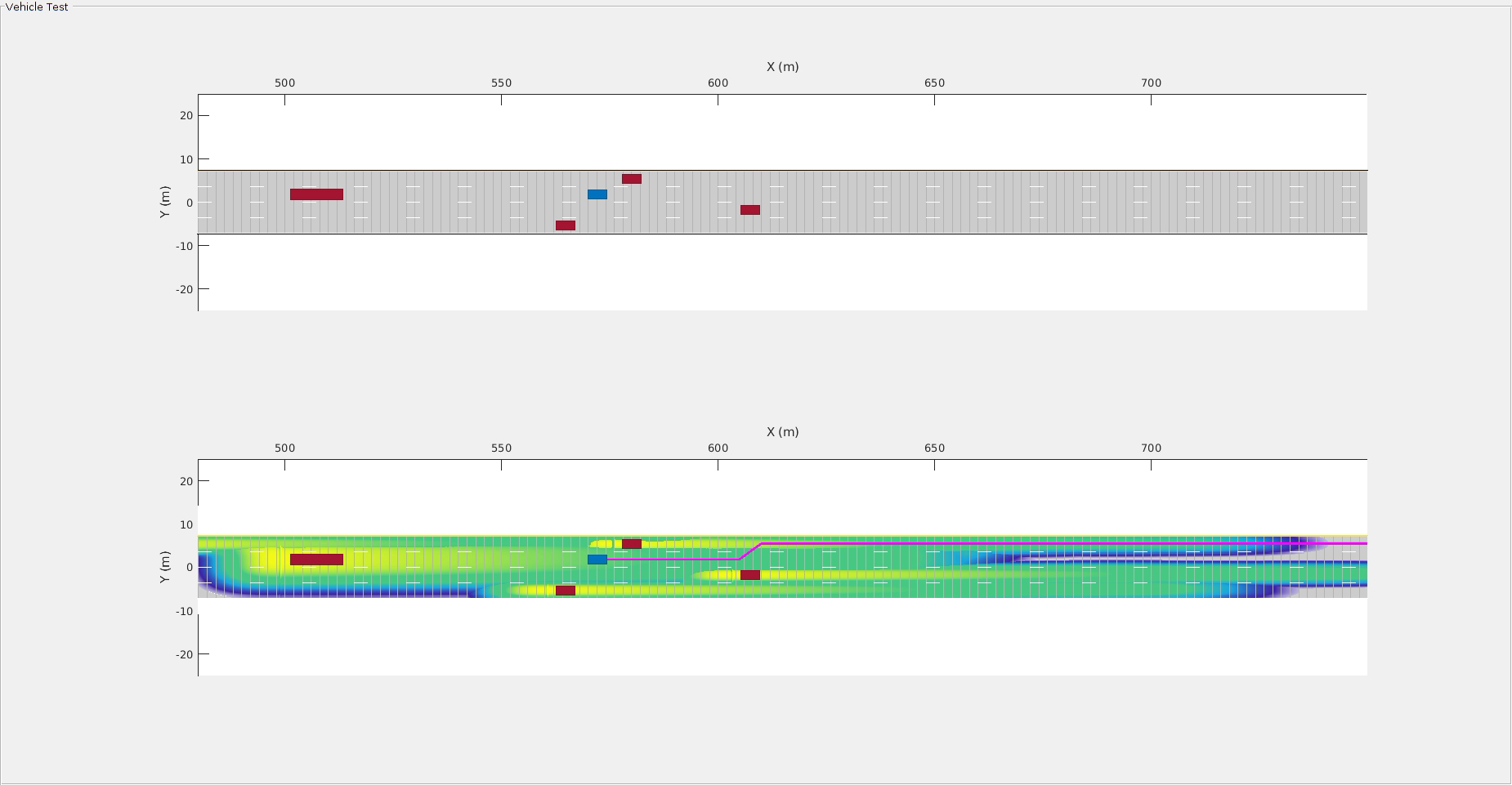}}

\caption{A demonstration of the model's ability to weave through traffic. The ego vehicle starts behind all vehicles on the road, but with a higher velocity and risk threshold $\mathcal{H}_{P}$ than all surrounding vehicles. The risk-aware planner navigates a safe human-like route to weave through the traffic. \href{https://youtu.be/El7k68taNEY}{\color{blue}{Video}}}
\label{fig_traffic_weave} 
\end{figure}

In \autoref{fig_traffic_weave}, the ego vehicle, with a higher desired velocity and risk threshold $\mathcal{H}_{P}$, displays more aggressive behavior compared to its surroundings, weaving through traffic to emulate aggressive driving. Current human behavior models struggle in weaving through dense traffic due to two main limitations. Firstly, models with short planning horizons lack sufficient time to plan a path through dense traffic, causing the ego vehicle to follow dense traffic groups rather than weaving through them. Secondly, when existing models attempt weaving, they frequently collide with surrounding vehicles due to a lack of consideration for the multimodal nature of surrounding vehicle trajectories which leads to high levels of uncertainty in the actions of other vehicles. Our approach, with its long planning horizon and multimodal trajectory prediction, safely navigates traffic, accurately representing aggressive driving behavior.

\subsubsection{Moving out of way of high-speed vehicle}

\begin{figure}[htbp]
\deemph{t = 0.0s} \\
\centerline{\includegraphics[scale=0.435, trim={11cm 8.1cm 35cm 23cm}, clip]{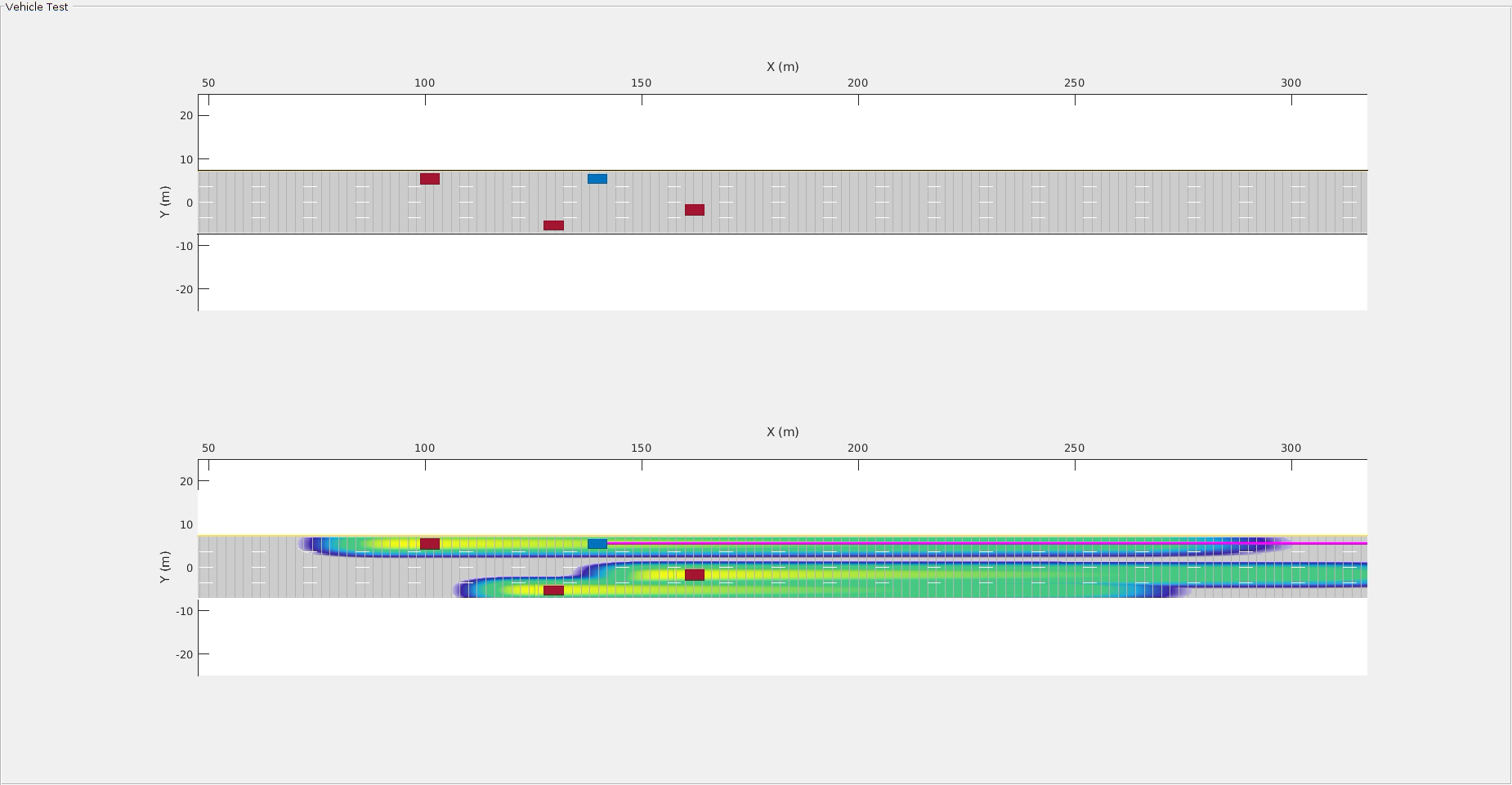}}
\deemph{t = 0.4s} \\
\centerline{\includegraphics[scale=0.435, trim={11cm 8.1cm 35cm 23cm}, clip]{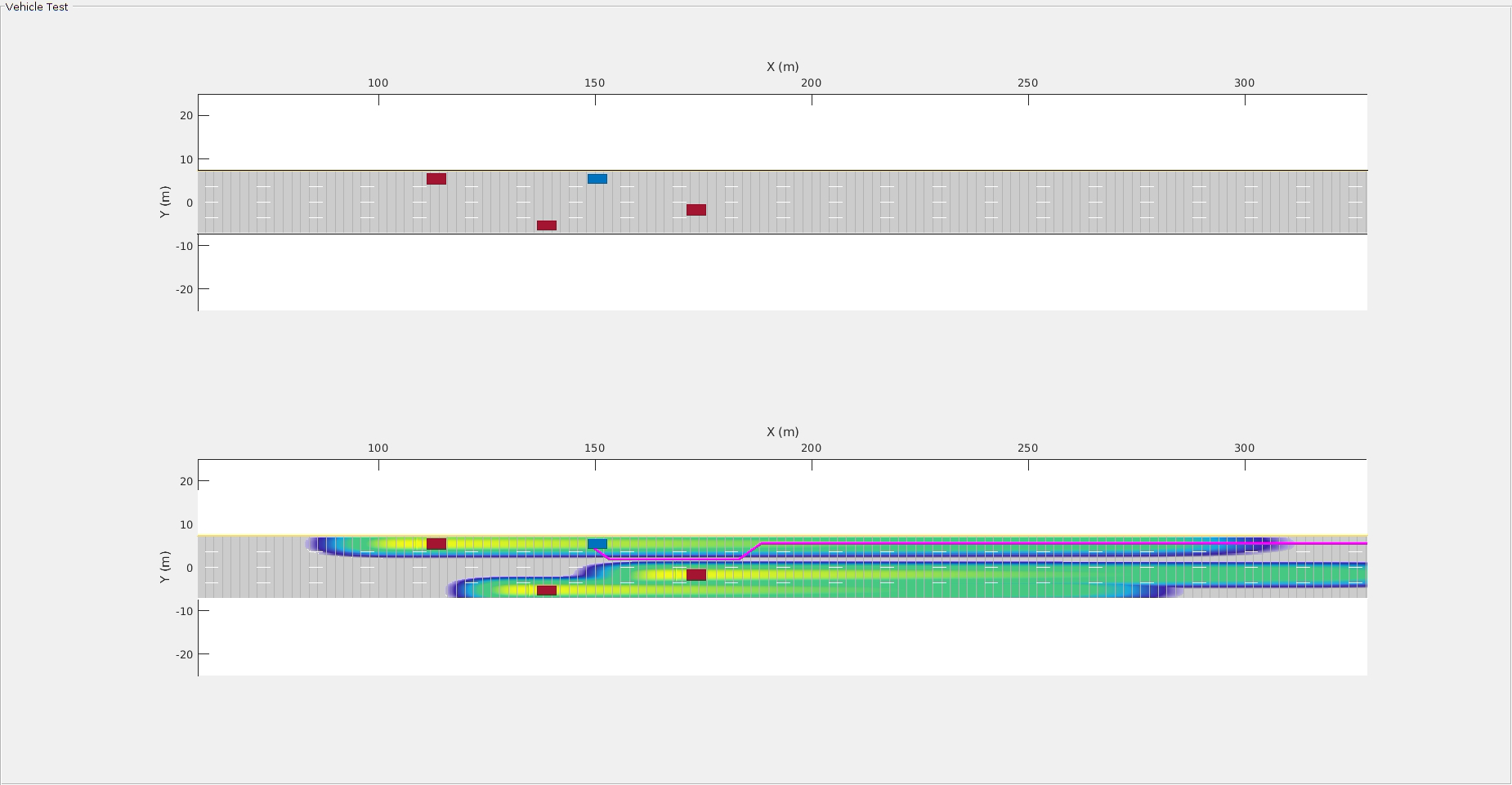}}
\deemph{t = 2.0s} \\
\centerline{\includegraphics[scale=0.435, trim={11cm 8.1cm 35cm 23cm}, clip]{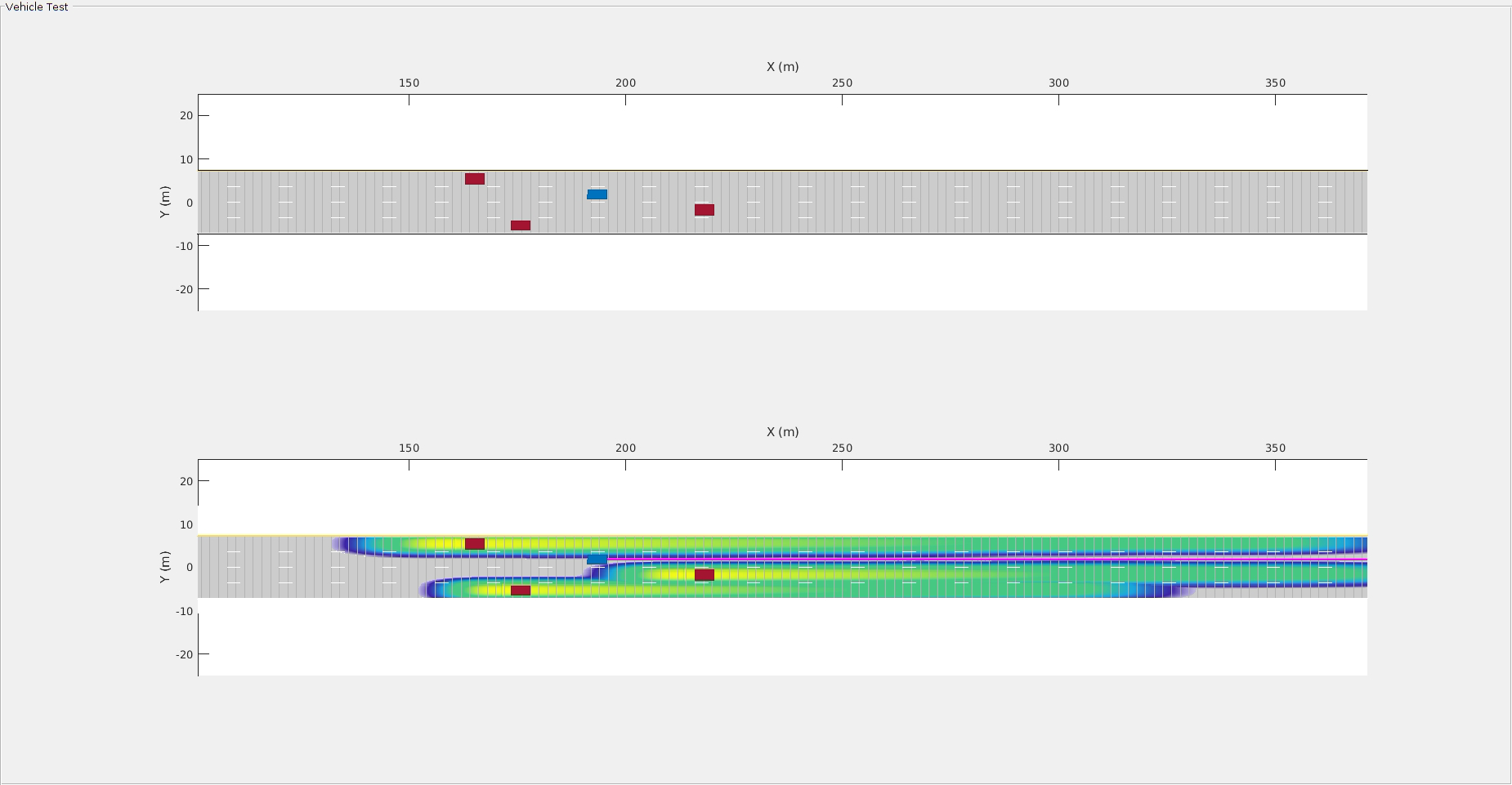}}
\deemph{t = 6.4s} \\
\centerline{\includegraphics[scale=0.435, trim={11cm 8.1cm 35cm 23cm}, clip]{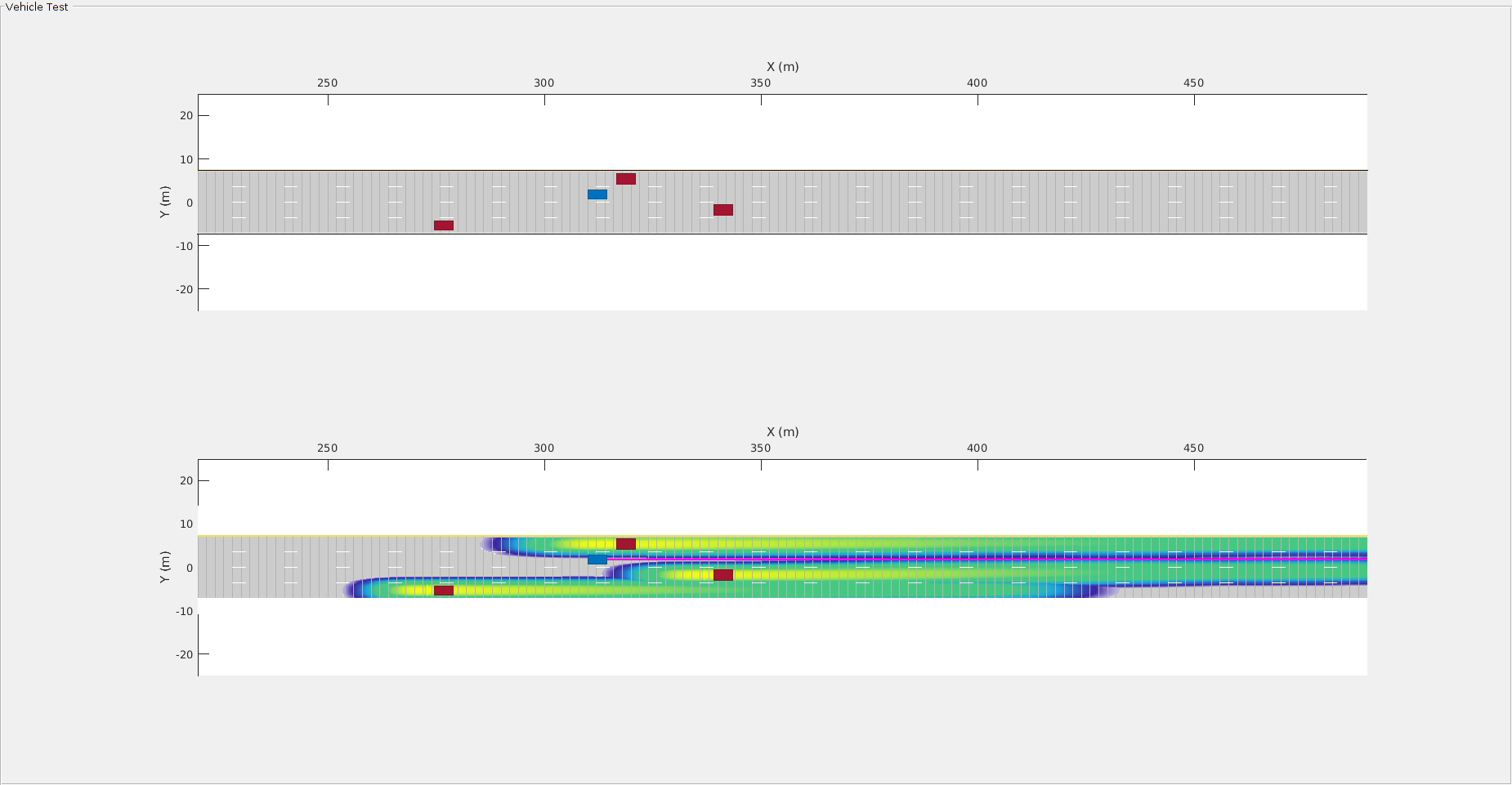}}
\caption{A demonstration of the model's ability to move out of the way of an approaching high-speed vehicle to maintain safety. The ego vehicle can be seen starting in front of a high-speed vehicle and traveling at a slower velocity and risk threshold $\mathcal{H}_{P}$ than the approaching vehicle. The planner navigates a safe human-like route to move out of the way. \href{https://youtu.be/SjX5KsTUTF8}{\color{blue}{Video}}}
\label{fig_move_out_of_way}
\end{figure}

In \autoref{fig_move_out_of_way}, the ego vehicle, with a lower desired velocity and risk threshold $\mathcal{H}_{P}$, exhibits a more cautious approach than surrounding vehicles as it encounters a fast-approaching car from behind. Our Risk-Aware planning framework guides the ego vehicle to change lanes, allowing the faster vehicle to pass. This behavior accurately models a conservative driver prioritizing safe driving by making way for a faster driver. Most current driver models lack this behavior due to inadequate forward-looking risk assessments for surrounding vehicles. Consequently, they fail to predict potential risks from other vehicles early enough, resulting in a keep-lane trajectory that causes the following vehicle to decelerate, even when a conservative human driver would change lanes. With our forward-looking risk assessment, the ego vehicle predicts the risk posed by the approaching vehicle well in advance, enabling it to yield the lane to the faster vehicle, mirroring the behavior of a cautious human driver.

\section{Conclusion}

In this paper, we propose the hierarchical learned risk-aware planning framework for modeling human drivers and validate and demonstrate its effectiveness in replicating human behavior. We additionally showcase the model's ability to generate human-like driving patterns in diverse, unseen situations. Future work will involve enhancing the model to encompass different driving characteristics~\cite{lyu2021probabilistic,lyu2022responsibility}, such as distracted driving, broadening the range of drivers it can represent. We also plan to use methods like probing~\cite{wang_active_2023} to actively learn additional parameters to improve the models ability to mimic human behavior across a wider array of situations and conduct further validation in diverse contexts to confirm its generalizability to other driving scenarios. 

\newpage

\printbibliography

\end{document}